\documentclass{article}

\usepackage{microtype}
\usepackage{graphicx}
\usepackage{subfigure}
\usepackage{booktabs} 
\usepackage{algorithm2e}
\usepackage{amsfonts}       
\usepackage{nicefrac}       
\usepackage{microtype}      
\usepackage{multirow}
\usepackage{graphicx}
\usepackage{svg}
\usepackage{bm}
\usepackage[moderate]{savetrees}

\usepackage{hyperref}

\usepackage[accepted]{icmlw2019generalization}

\icmltitlerunning{Zero-Shot Learning from scratch (ZFS): leveraging local compositional representations}

\begin{document}

\twocolumn[
\icmltitle{Zero-Shot Learning from scratch (ZFS): leveraging local compositional representations}



\icmlsetsymbol{equal}{*}

\begin{icmlauthorlist}
\icmlauthor{Tristan Sylvain}{equal,ml}
\icmlauthor{Linda Petrini}{equal,ed}
\icmlauthor{R Devon Hjelm}{ms,ml}
\end{icmlauthorlist}

\icmlaffiliation{ml}{Mila, University of Montreal, Montreal, Canada}
\icmlaffiliation{ms}{Microsoft Research, Montreal, Canada}
\icmlaffiliation{ed}{University of Amsterdam, Amsterdam, Netherlands}

\icmlcorrespondingauthor{Tristan Sylvain}{tristan.sylvain@gmail.com}

\icmlkeywords{Machine Learning, ICML}

\vskip 0.3in
]



\printAffiliationsAndNotice{\icmlEqualContribution} 

\begin{abstract}
Zero-shot classification is a generalization task where no instance from the target classes is seen during training. 
To allow for test-time transfer, each class is annotated with semantic information, commonly in the form of attributes or text descriptions.
While classical zero-shot learning does not explicitly forbid using information from other datasets, the approaches that achieve the best absolute performance on image benchmarks rely on features extracted from encoders pretrained on Imagenet. 
This approach relies on hyper-optimized Imagenet-relevant parameters from the supervised classification setting, entangling important questions about the suitability of those parameters and how they were learned with more fundamental questions about representation learning and generalization.
To remove these distractors, we propose a more challenging setting: Zero-Shot Learning from scratch (ZFS), which explicitly forbids the use of encoders fine-tuned on other datasets. Our analysis on this setting highlights the importance of local information, and compositional representations.
\end{abstract}

\section{Introduction}
Zero-Shot Learning (ZSL)~\citep{larochelle2008zero} is a difficult classification framework introduced to push training, understanding, and evaluating models towards those whose performance \emph{generalizes} to new, unseen concepts. 
In order to perform well under this framework, learned models must be able to make ``useful" inferences about unseen concepts (e.g., correctly label), given parameters learned only from seen training concepts and additional semantic information.
This is fundamentally an important framework for evaluating and understanding models meant to be used in real world scenarios, where a representative sample of labeled training data is expensive, and/or not all relevant test cases may be known at training time.
Training models that perform well under this framework requires thinking beyond the normal classification by incorporating ideas such as systematic generalization~\citep{bahdanau2018systematic} and compositionality~\citep{tokmakov2018learning}.

While the original ZSL setting introduced in \citet{larochelle2008zero} was agnostic to the exact methodology, more recent image ZSL approaches almost uniformly use features from very large ``backbone" neural networks such as InceptionV2~\citep{szegedy2016rethinking} and ResNet101~\citep{he2016deep} pretrained on the Imagenet dataset~\citep{ILSVRC15}.
In terms of absolute performance, this approach appears to be well-justified, as state-of-the-art results on various ZSL~\citep{yosinski2014transferable, sun2017revisiting, huh2016makes, azizpour2015factors} and non-ZSL benchmarks~\citep{li2019analysis, zhang2018fine, he2017fine} all learn on top of these pretrained backbones.
There is also a valid, intuitive analogy between pretraining on large datasets to how humans are exposed to large amounts of diverse data during their ``training"~\citep{biederman1987recognition}.

However, we have many concerns with this approach.
First, relative success in transfer learning has been shown to be highly dependent on the pretrained backbone encoder~\citep{xian2018zero} on Imagenet.
In addition, using different datasets in the pretraining phase has been shown to lead to very different transfer learning performance on related tasks~\citep{cui2018large}.
Next, while Imagenet features have been shown to work well for ZSL tasks with similar image datasets, there are no guarantees that such a pretraining framework would exist with general ZSL settings, either in existence or suitability for the task. 
Conversely, it can be hard in practice to meaningfully evaluate a Zero-Shot learner, as some of the benchmarking test classes have been shown to be present in the pretraining dataset~\citep{DBLP:journals/corr/XianLSA17}.

Finally, we believe this approach misses the point. 
ZSL isn't meant to be primarily an exercise of achieving the best absolute performance on a set of benchmarks; it should first and foremost be used as a framework for training, understanding, and evaluating models on their ability to reason about new, unseen concepts. 
Despite the \emph{absolute performance} gains of the methods above that use Imagenet features, the use of backbones hyper-optimized for supervised performance on Imagenet and the Imagenet dataset itself represent \emph{nuisance variables} in a larger effort to understand how to learn generalizable concepts from scratch.

For these reasons and more, we wish to return ZSL back to its roots, distinguishing it from the bulk of recent ZSL literature, by ``introducing"~\emph{Zero-Shot Learning from scratch} (ZFS).\footnote{This important framework isn't really new, but it has been ignored by the community, so we feel it is appropriate to give it a new name.}
In this setting, the model is evaluated on its ability to perform classification using auxiliary attributes and labels trained only using the data available from the training split of the target dataset.
We believe that ZFS will provide researchers with a better experimental framework to understand the factors that improve performance in Zero-Shot generalization.


ZFS is a fundamentally harder setting than that of using encoders pretrained on Imagenet, as we need to learn all lower-level features from scratch in a way that will generalize to unseen concepts.
As a first good step to solving ZFS, we follow intuitions from works on compositionality and locality~\citep{tokmakov2018learning, stone2017teaching}, focusing on methods that encourage good local representations.

The contributions of our work are as follows:
\begin{itemize}
    \item We introduce Zero-Shot Learning from scratch (ZFS), an extention of the ZSL task, which we believe will be an important benchmark for understanding the generalization properties of learning algorithms, models, and encoder backbones.
    \item We evaluate several supervised and unsupervised methods on their ability to learn features that generalize in the ZFS setting by training a prototypical network on top of those features \citep[in a similar way to what was done in][with Imagenet features]{snell2017prototypical}.
    \item Motivated by compositionality and locality, we introduce a local objective to the above methods and show improvement on the ZFS task.
\end{itemize}

\begin{figure}[ht]
  \centering
  \includegraphics[width=0.45\textwidth]{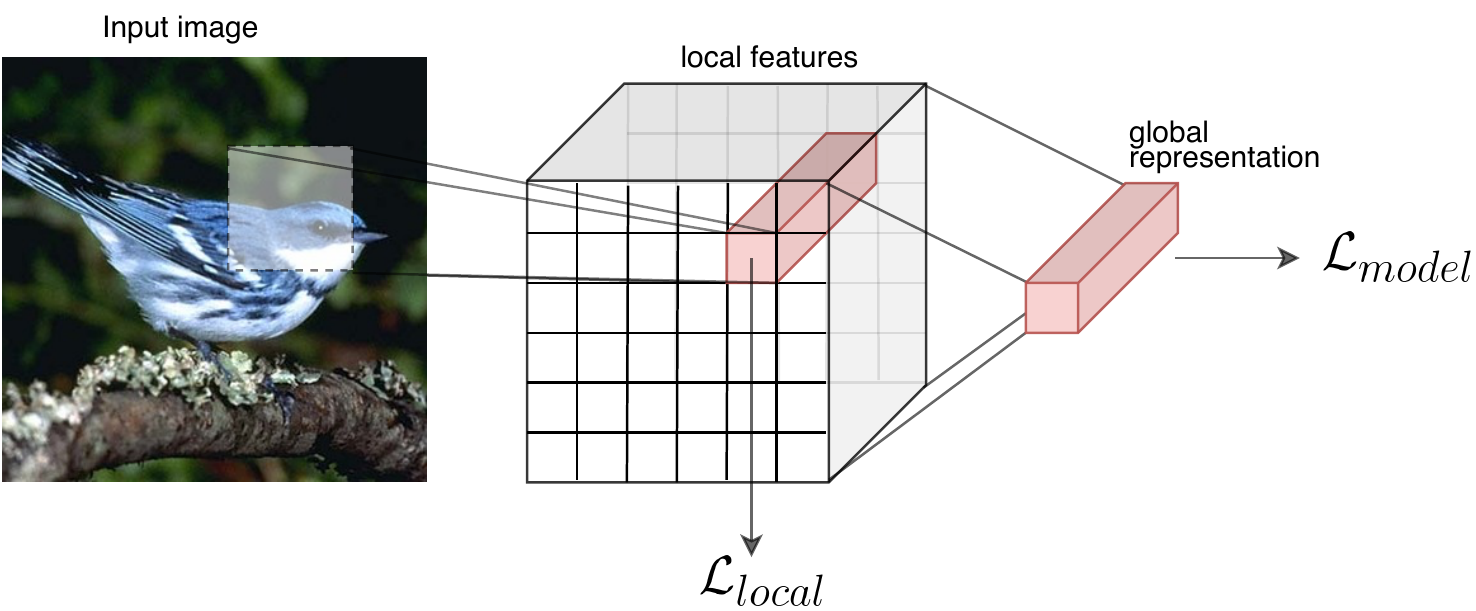}
  \caption{\textbf{Local objective}. The encoder takes as input an image and outputs a global representation - used to compute the model loss $\mathcal{L}_{model}$. To encourage locality and compositionality, label or attribute based classification at earlier layers in the encoder network ($\mathcal{L}_{local}$) is performed.}
  \label{fig:local_objective}
\end{figure}

\section{Zero-Shot Learning from scratch}
\subsection{Definition}
Following our concerns outlined above with performing ZSL using pretrained Imagenet encoders, we introduce Zero-Shot Learning from Scratch (ZFS), which we believe provides a better experimental framework to understand the factors that contribute to generalization to unseen concepts at test-time.
In addition to the ZSL experimental framework outlined in \citet{larochelle2008zero}, ZFS simply adds one additional additional requirement:

\textbf{ZFS requirement: }\emph{No model parameters can contain information about (e.g., can be learned from) data outside that from the training split of the target dataset.}

For example, for the Caltech-UCSD-Birds-200-2011~\citep[CUB,][]{wah2011caltech} dataset, the model to be evaluated under ZFS can only be learned using sample birds the training split. 
ZFS removes nuisance variables related to pretraining from understanding \emph{why} different learning algorithms might work, thus making comparisons more meaningful.
ZFS also opens to door to using encoder backbones that are non-standard and \emph{not hyper-optimized} for a different task, the suitability of which is not well understood or studied in the Imagenet-pretrained setting.  

\subsection{Hypothesis and Approach}
For images, we hypothesize that good performance in ZFS requires representations that are \emph{locally compositional}.
For example in the CUB dataset, locally compositional would mean that the bird representations is a composition of local ``part" representations (e.g., representations of ``yellow beak", ``black wings", etc).
We and others~\citep{zhu2019parts} believe that learning representations this way in this setting will generalize well, as samples in the train and test split have similar marginal distributions over the parts, yet vary in their composition (i.e., in their joint distributions). 
Once good part representations are learned, a typical ZSL method such as prototypical networks~\citep{snell2017prototypical} can be used to classify by matching their composition with given class-specific semantic features.

In this work, we train convolutional image encoders using either supervised or unsupervised learning, then use prototypical networks to perform zero-shot learning onto these fixed representations. 
Using prototypical networks for ZFS requires minimal parameters or hyper-parameters tuning, is well-studied~\citep{huang2019centroid, finn2017model}, and performance is very close to the state of the art for Imagenet-pretrained benchmarks.
We encourage the image encoder to extract semantically relevant features at earlier stages of the network 
by introducing an auxiliary local loss to the local feature vectors of a convolutional layer (which corresponds to a patch patch of the image).

A visual representation of our approach is provided in Figure~\ref{fig:local_objective}, while the training procedure is described in Algorithm~\ref{alg:training_alg}. We introduce one of two additional auxiliary classification tasks onto the local features of one (or more) of the earlier layers of the encoder. 
The first auxiliary task - which we refer to as Local Label-based Classifier (\textbf{LC}) - consists of an additional simple classifier, trained with a standard cross entropy loss over the training classes. 
The second auxiliary task - Local Attribute-based Classifier (\textbf{AC}) - trains a separate prototypical network to project local features to an embedding space that matches attribute embeddings. 
The local loss $\mathcal{L}_{local}$ is then added with the model's base loss $\mathcal{L}_{model}$ (e.g. reconstruction loss for VAEs, discriminator loss for AAE) and all models are trained end-to-end using back-propagation and stochastic gradient descent.
\vspace{-0.3cm}


\begin{algorithm}[H]
\SetAlgoLined
\textbf{Input:} Training set $\mathcal{D}$,
Encoder $E_{\phi}(\bm{x})$, Model $M_{\phi}(\bm{g})$, Local Model $L_{\phi}(\bm{l})$, Prototypical Network $P_{\theta}(\bm{g}, \bm{a})$.

\textbf{Output:} Parameters $\phi$ and $\theta$ after a training epoch.

\For{ batch $i\leftarrow 1$ \KwTo $N$}{
    local features $\bm{l}_i$, global features $\bm{g}_i = E(\bm{x}_i)$ \\
    $\mathcal{L}_{local} =\textit{loss}(L(\bm{l}_i, \bm{a}_i), y_i$) \\
    $\mathcal{L}_{global} = \textit{loss} (M(\bm{g}_i), y_i, \bm{x}_i)$ \\
    $J =  \mathcal{L}_{local} + \mathcal{L}_{global}$ \\
    Update $E_{\phi}, M_{\phi}, L_{\phi}$ with $\nabla_{\phi} J(\phi) $ \\
    A = \textit{loss} ($P(\bm{g}_i$\texttt{.detach()}, $\bm{a}_i ), y_i$ ) \\
    Update $P_{\theta}$ with $\nabla_{\theta} A(\theta) $
 }
\caption{Training procedure.}
\label{alg:training_alg}
\end{algorithm}

\section{Related works}
\subsection{Zero-Shot Learning}

Zero-Shot Learning (ZSL) can be formulated as a meta-learning problem~\citep{vinyals2016matching}. The absence of examples from the test distribution to fine-tune the model (contrary to Few-Shot Learning~\citep[FSL,][]{li2006one}) results in the fact that methods that learn an initialization such as MAML~\citep{finn2017model} are hard to apply, and almost all recent state-of-the-art methods~\citep{Akata_2015, Changpinyo_2016, Kodirov_2017, Zhang_2017, Sung_2018} rely instead on metric learning by applying two steps: (1) learning a suitable embedding function that maps data samples and class attributes to a common subspace, (2) performing nearest neighbor classification at test-time with respect to the embedded class attributes.

\subsection{Compositionality}
Compositional representations have been a focus in the cognitive science literature~\citep{biederman1987recognition, hoffman1984parts} with regards to the ability of intelligent agents to generalize to new concepts. 
There has recently been a renewed interest in learning compositional representations for applications such as computational linguistics~\citep{tian2016learning}, and generative models~\citep{higgins2017scan}. 
For FSL, \citet{tokmakov2018learning} introduced an explicit penalty to encourage compositionality when dealing with attribute representations of classes. \citet{alet2018modular} encourages compositionality in meta-learning by using a modular architecture. In this work, we focus on implicit constraints to encourage compositionality, adding a local attribute or label classifier to provide additional local gradients to the encoder.

\subsection{Locality}
Local features have been extensively used to improve different model's performance. Self-attention over local features resulted in large improvements in generative models~\citep{zhang2018self}. Attention over local features is commonly used in image captioning~\citep{li2017image}, visual question answering~\citep{kim2018bilinear} and fine-grained classification~\citep{sun2018multi}.

Deep InfoMax~\citep[DIM,][]{hjelm2018learning} is an unsupervised model that maximizes the mutual information maximization between local and global features to learn unsupervised representations. 
Performance of DIM is promising for unsupervised representation learning using linear or nonlinear classification probes, and extensions have achieved state-of-the-art on numerous related tasks~\citep{velivckovic2018deep, bachman2019learning}.
Because of this, we hypothesize that DIM may learn good representations suitable for ZFS, and this is validated in our experiments.

\section{Experiments}
\subsection{Setup}


\paragraph{Datasets}
We use the following standard ZSL benchmarks for ZFS: Animals with Attributes 2~\citep[\textbf{AwA2},][]{xian2018zero}, 30,475 images, 50 different animal classes, attributes of dimension 85. Caltech-UCSD-Birds-200-2011~\citep[\textbf{CUB},][]{wah2011caltech}, 11,788 images, 200 different types of birds, attributes of dimension 312. SUN Attribute~\citep[\textbf{SUN},][]{Patterson2012SunAttributes}, 14,340 images, 717 types of scenes, attributes of dimension 102. As in ~\citep{DBLP:journals/corr/XianLSA17} we used $\ell_2$ normalized versions of these attributes as semantic class representations. 

\paragraph{Compared methods}
We perform comparisons between a set of common representation learning methods, both supervised and unsupervised. All methods are trained using only the data in the train set, which includes images, class labels, and class attributes. The methods selected for comparison are: Fully supervised label classifier (\textbf{FC}), Variational auto-encoders (\textbf{VAE})~, $\beta$ Variational auto-encoders (\textbf{$\beta$-VAE}), Adversarial auto-encoders (\textbf{AAE}) and Deep InfoMax (\textbf{DIM}).

\begin{table*}[ht]
\centering
\caption{\textbf{Zero-shot accuracy of the different models.} LC stands for a local label-based classifier trained with cross-entropy. AC stands for a local attribute-based classifier similar to prototypical networks ($\bullet$ denotes presence, $\circ$ absence). Alex and Basic refer to whether AlexNet or a basic CNN is used as the encoder. Bold is best result per model across the different kind of tasks and encoder architectures. Results show how encouraging locality improves accuracy.}
\begin{tabular}{|c||c||c|c||c|c||c|c|}
\hline
\multicolumn{1}{|c||}{\multirow{2}{*}{Model}} & \multicolumn{1}{c||}{Aux classifiers} & \multicolumn{2}{c||}{CUB} & \multicolumn{2}{c||}{AwA2} & \multicolumn{2}{c|}{SUN} \\ \cline{3-8} 
\multicolumn{1}{|c||}{}                    & \multicolumn{1}{c||}{LC \thinspace AC}                     & Alex       & Basic       & Alex       & Basic       & Alex        & Basic       \\ \hline \hline
    \multirow{1}{*}{FC}  & \begin{tabular}[t]{@{}cc@{}}\phantom{h}$\circ$ & $\circ$ \end{tabular} & \textbf{32.09} & 25.64 & 37.37 & \textbf{37.97} & \textbf{23.57} & 19.50  \\ \hline \hline
    \multirow{1}{*}{Prototypical Networks \cite{snell2017prototypical}}  & \begin{tabular}[t]{@{}cc@{}}\phantom{h}$\circ$ & $\circ$ \end{tabular} & 33.56 & 28.57 & 37.37 & 37.21 & 23.13 & 23.06 \\ \hline \hline
   \multirow{3}{*}{VAE~\citep{kindgma2013vae}}  & \begin{tabular}[t]{@{}cc@{}}\phantom{h}$\bullet$ & $\circ$ \end{tabular} & 14.71 & \textbf{15.08} & \textbf{30.59} & 30.55 & 15.20 &  \textbf{16.12}  \\      
                       \cline{2-8} & \begin{tabular}[t]{@{}cc@{}}\phantom{h}$\circ$ & $\bullet$ \end{tabular} & 13.61 & 14.44 & 29.70 & 30.50 & 14.86 & 15.42  \\
                       \cline{2-8} & \begin{tabular}[t]{@{}cc@{}}\phantom{h}$\circ$ & $\circ$ \end{tabular} & 13.78 & 14.52 & 29.71 & 30.39 & 14.61 & 15.08  \\ \hline \hline
   \multirow{3}{*}{$\beta$-VAE~\citep{Higgins2017betaVAELB}}  & \begin{tabular}[t]{@{}cc@{}}\phantom{h}$\bullet$ & $\circ$ \end{tabular} & 14.16 & \textbf{15.25} & 30.16 & \textbf{30.75} & 14.77 &  \textbf{16.12}  \\      
                       \cline{2-8} & \begin{tabular}[t]{@{}cc@{}}\phantom{h}$\circ$ & $\bullet$ \end{tabular} & 13.07 & 13.34 & 29.34 & 29.83 & 14.38 & 14.93  \\
                       \cline{2-8} & \begin{tabular}[t]{@{}cc@{}}\phantom{h}$\circ$ & $\circ$ \end{tabular} & 13.91 & 14.05 & 30.10 & 29.38 & 14.67 & 14.88  \\ \hline \hline
   \multirow{3}{*}{AAE~\citep{makhzani2015aae}}  & \begin{tabular}[t]{@{}cc@{}}\phantom{h}$\bullet$ & $\circ$ \end{tabular} & 17.46 & \textbf{18.14} & 30.56 & 31.57 & 17.47 & 16.83   \\      
                       \cline{2-8} & \begin{tabular}[t]{@{}cc@{}}\phantom{h}$\circ$ & $\bullet$ \end{tabular} & 16.68 & 18.00 & 30.31 & 30.98 & 16.67 & \textbf{17.57}  \\
                       \cline{2-8} & \begin{tabular}[t]{@{}cc@{}}\phantom{h}$\circ$ & $\circ$ \end{tabular} & 17.24 & 17.76 & 29.71 & \textbf{31.76} & 16.92 & 16.78  \\ \hline \hline
   \multirow{3}{*}{DIM~\citep{hjelm2018learning}}  & \begin{tabular}[t]{@{}cc@{}}\phantom{h}$\bullet$ & $\circ$ \end{tabular} & 32.37 & 28.46 & 41.86 & 42.55 & 32.78 & 29.44   \\      
                       \cline{2-8} & \begin{tabular}[t]{@{}cc@{}}\phantom{h}$\circ$ & $\bullet$ \end{tabular} & \textbf{32.45} & 30.26 & 42.45 & \textbf{43.62} & 34.38* & \textbf{35.35}  \\
                       \cline{2-8} & \begin{tabular}[t]{@{}cc@{}}\phantom{h}$\circ$ & $\circ$ \end{tabular} & 29.56 & 28.25 & 40.72 & 39.48 & 31.94 & 32.71  \\ \hline 

\end{tabular}
\label{tab:results}
\end{table*}

\paragraph{Image encoders}
While common Zero-Shot Learning methods consider large encoders pretrained on Imagenet~\citep{ILSVRC15} to simplify experiments and due to capacity constraints, we choose to consider only smaller networks. We considered both a small encoder derived from the DCGAN architecture~\citep{radford2015unsupervised} and similar in capacity to those used in early few-shot learning models such as MAML~\citep{finn2017model}. We also consider AlexNet~\citep{krizhevsky2012Imagenet} to gain insight on the impact of the encoder backbone. 
It is important to note that overall the encoders we use are significantly smaller than the ``standard" backbones common in state-of-the-art Imagenet-pretrained ZSL methods. We believe restricting the encoder's capacity decreases the overall complexity, but does not hinder our ability to extract understanding of what methods work from our experiments.


\paragraph{Evaluation Protocol}
To perform a rigorous evaluation of zero-shot classification tasks, the sets of train and test classes need to be strictly disjoint.
\citet{DBLP:journals/corr/XianLSA17} shows how the latter does not hold for the most commonly used train/test splits in previous ZSL work, due to the presence of many test examples in the Imagenet training set. 
They propose a new data split addressing these issues, the Proposed Split (\textbf{PS}), which we use in our experiments.
The number of train/test classes is: CUB - 150/50, AwA2 - 40/10, SUN 645/72.
All models are evaluated on Top-1 accuracy.
We pretrain the encoder using each of the previously mentioned methods (strictly on the considered dataset, as per the ZSF requirement). We then train a prototypical network on top of the (fixed) learned representation. For each case, we consider the effect of adding local label-based classifiers (\textbf{LC}) and local attribute-based classifiers (\textbf{AC}) during the encoder training. 
\vspace{-0.4cm}


\paragraph{Implementation details}
All models used in this paper have been implemented in PyTorch. We use a batch size of 64. All images have been resized to size $128\times128$. During training, random crops with aspect ratio $0.875$ were performed. During test, center crops with the same ratio were used. While most ZSL approaches do not use crops (due to the fact that they used pre-computed features), this experimental setup was show to be efficient in the field of text to image synthesis~\citep{reed2016generative}. All models are optimized with Adam and a learning rate of 0.0001. The final output of the encoder is of dimension 1024 across all models. Local experiments were performed extracting features from the third layer in the network. These features have dimension $27\times27\times384$ for the AlexNet based encoder and $14\times14\times256$ for the simple CNN encoder. 

\subsection{Results}
Results of our proposed approach in the framework of ZFS are displayed in Table~\ref{tab:results}. 
For all methods considered (with the exception of AAE on AwA2), the addition of local information results in an increase in Top-1 accuracy. For the models whose loss is only based on the global representation (VAE, $\beta$-VAE, AAE), the label-based local task performs better than the attribute based one, but not by a significant margin. 

The best performing model is DIM, whose accuracy is comparable to (on CUB) and significantly higher (on AwA2 and SUN) than the fully supervised model (FC). Moreover, DIM is the only model for which the attribute-based local task shows significant improvement over the label-based one. DIM is by definition a model that strongly leverages on local information, supporting our hypothesis that locality is a fundamental ingredient for generalization. 

These results suggest that the proposed auxiliary losses can have a significant positive influence on models who already have a notion of locality, but are less effective in the case where the model mostly relies on global information (as a reconstruction task necessarily needs to take into account the whole input).

\vspace{-0.2cm}
\section{Conclusion and future work}
Motivated by the need for more realistic evaluation settings for Zero-Shot Learning methods, we proposed a new evaluation framework where training is strictly performed only on the benchmark data, with no pretraining on additional datasets. In the proposed setting, we hypothesize that the fundamental ingredients for successful transfer learning are locality and compositionality. We propose an auxiliary loss term that encourages these characteristics and evaluate a range of models on the CUB, AwA2 and SUN datasets. We observe how the proposed approach yields to improvements in ZSL accuracy, thus confirming our hypothesis.

\bibliography{bibliography}
\bibliographystyle{icml2019}

\end{document}